%% file: main.tex
\documentclass[letterpaper, 10 pt, journal, twoside]{IEEEtran}     

\usepackage{graphicx}
\usepackage{algorithm}
\usepackage[noend]{algpseudocode}
\usepackage{amsmath}
\usepackage{nicematrix} 
\usepackage{multirow}  
\usepackage{comment}
\usepackage{float} 

\usepackage{tikz, pgfplots, fix-cm}
\usetikzlibrary{plotmarks}
\usepackage{graphicx,booktabs,array}

\usepackage{amssymb}
\usepackage{pifont}
\usepackage{efbox}
\efboxsetup{linecolor=black,linewidth=1pt}

\def\secref#1{Sec.~\ref{#1}}
\def\figref#1{Fig.~\ref{#1}}
\def\tabref#1{Tab.~\ref{#1}}
\def\eqref#1{Eq.~(\ref{#1})}

\newcommand{\RNum}[1]{(\uppercase\expandafter{\romannumeral #1\relax})}

\usepackage{amsmath}
\usepackage{xcolor}

\newcommand\etal{\emph{et al.}}
\newcommand\bbot{BonnBot-I}

\def\argmax{\mathop{\rm argmax}}

\title{\LARGE \bf BonnBot-I Plus: A Bio-diversity Aware Precise Weed Management Robotic Platform}

\author{Alireza Ahmadi$^{1}$, Michael Halstead$^{1}$, Claus Smitt$^{1}$, and Chris McCool$^{1,2}$
	\thanks{$^{1}$University of Bonn, Bonn 53115 Germany. 
			{\tt\small \{alireza.ahmadi, michael.halstead, csmitt, cmccool\}@uni-bonn.de}}%
	\thanks{$^{2}$Lamarr Institute for Machine Learning and Artificial Intelligence, \url{lamarr-institute.org}}%
}

\begin{document}

\maketitle
\thispagestyle{empty}
\pagestyle{empty}

\begin{abstract}
In this article, we focus on the critical tasks of plant protection in arable farms, addressing a modern challenge in agriculture: integrating ecological considerations into the operational strategy of precision weeding robots like \bbot. 
This article presents the recent advancements in weed management algorithms and the real-world performance of \bbot\ at the University of Bonn's Klein-Altendorf campus.
We present a novel Rolling-view observation model for the BonnBot-Is weed monitoring section which leads to an average absolute weeding performance enhancement of $3.4\%$.
Furthermore, for the first time, we show how precision weeding robots could consider bio-diversity-aware concerns in challenging weeding scenarios.
We carried out comprehensive weeding experiments in sugar-beet fields, covering both weed-only and mixed crop-weed situations, and introduced a new dataset compatible with precision weeding.
Our real-field experiments revealed that our weeding approach is capable of handling diverse weed distributions, with a minimal loss of only $11.66\%$ attributable to intervention planning and $14.7\%$ to vision system limitations highlighting required improvements of the vision system.

\end{abstract}

\section{Introduction}
Modern agriculture and weed control aims to support plant growth while considering environmental protection~\cite{zingsheim2024weeding}.
Weed control is a topical example of this because leaving weeds in a field can significantly impact crop yields as the weeds will compete, or even out-compete, the crops for vital nutrients and resources leading to potential reductions in productivity.
Yet, in recent years it has become clear that we need to be able to strike a balance between protecting crops (e.g. removing weeds) and the environment, in particular reducing the usage of herbicides for weed control~\cite{heap2014global}. 

Robotics has emerged as a potential tool to better enable this trade-off, particularly for the task of weed management.
As early as 2002~\cite {steward2002distance} researchers have considered the potential for robotics to perform precise weed control at the individual plant level.
Since then, multiple methods to perform robotic weed management have been proposed including physical~\cite{Bawden17_1,chang2021mechanical}, chemical~\cite{wu2020robotic}, electrical~\cite{ascard200710}, and more recently laser-based~\cite{xiong2017development}.
An issue with many of the above solutions is that they either do not provide flexibility for different tools or their implements have a large footprint that does not enable plant-specific interaction.

\begin{figure}[t]
	\centering
        \vspace{2mm}
	\includegraphics[width=1\linewidth]{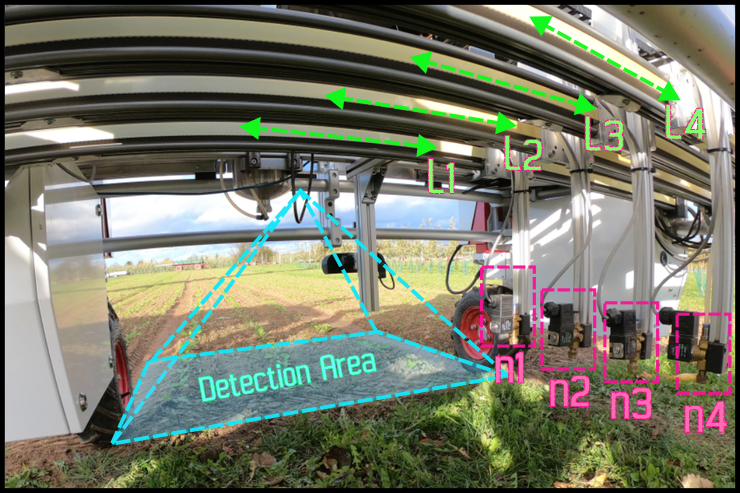}
	\caption{An image from underneath the \bbot\ Platform in sugar beet fields at Campus Klein Altendorf of the University of Bonn, prior to conducting Precision weed management. (n1-n4: Spray nozzles in purple, L1-L4: Linear axes in light-green, and the detection area of the front camera in cyan).}
	\label{fig:motivation}
        \vspace{-6mm}
\end{figure}

In this article, we outline developments on BonnBot-I~\cite{ahmadi2022bonnbot} which enhance its ability to perform bio-diversity-aware weed management.
In earlier work, we developed planning systems for BonnBot-I that moved its replicated linear system of tools to perform weed management, depicted in~\figref{fig:motivation}.
We propose an alternative planning approach that emphasizes the need for robots to recognize and adapt to the diversity of weed species.
By differentiating weeds based on their specific characteristics and competitive behaviors, the method enhances crop protection and resource optimization in weeding, signifying a shift towards more nature-conscious and efficient agricultural practices.
%
%
These advancements on BonnBot-I led to the following contributions:
\begin{enumerate}
    \item A bio-diversity aware weed management approach.
    \item A rolling view strategy that improves weeding.
    \item Real-world experiments of multi-nozzle weeding.
    \item Releasing a multi-weed arable farming sugar-beet dataset \textbf{SB21}.
\end{enumerate}

This extends our prior work~\cite{ahmadi2022bonnbot} in three key areas. 
First, we introduce an improved planning method utilizing a novel Rolling-view observation method, enhancing performance by $3.4\%$. 
Second, we introduce a bio-diversity-aware weed management system, a first in the field, and assess its real-world applicability. 
Third, we validate our system's effectiveness through real-world experiments for multi-nozzle weed management, demonstrating its potential to optimize agricultural practices and productivity.

\section{Related Works}
Weeds can significantly impact crop yields by competing with them for vital nutrients and resources, leading to potential reductions in productivity~\cite{oerke2006crop, zimdahl2007weed}.
Hence, in today's world of agriculture, the efficient management of weeds is vital when it comes to maximizing crop yields and ensuring global food security. 
Traditionally, farmers have heavily relied on manual labor or chemical herbicides for weed control~\cite{slaughter2008autonomous}. 
However, these methods often entail labor-intensive processes, consume significant time, and come with potential environmental repercussions~\cite{guglielmini2017competitive}.
Such uniform treatment approaches result in both soil and yield damage~\cite{dentika2021weeds}.
Therefore pushing society toward more eco-friendly approaches to strike a balance between protecting crops and the environment is particularly important.
This type of eco-friendly approach will result in the reduction of herbicides and pesticides for weed control.
This concept of robotic-vision based farming seeks to revolutionize farming by enabling precision treatment at the individual plant level~\cite{steward2002distance}.
Selective precision weeding presents a promising alternative to the conventional and often excessive use of herbicides for weed control~\cite{ahmadi2022bonnbot}.
The innovations provided by agricultural robotics reduce labor, prevent soil compaction, and utilize perception-based monitoring, which cuts down on the agrochemicals used, directly benefiting the yield~\cite{ahmadi2022bonnbot}.

Advancement in field monitoring and computer vision applied in agriculture is an enabling factor for precision weed management.
For example, Zhu~\etal\ deployed an attention-based YOLO architecture for better weed detection~\cite{zhu2024research}.
Halstead~\etal~\cite{halstead2021crop} developed a crop-agnostic architecture based on MaskRCNN~\cite{he2017mask} to enable monitoring in arable farms as well as glasshouses.
Furthermore, a wide range of weed management tools have been investigated including mechanical~\cite{Bawden17_1, chang2021mechanical}, chemical~\cite{wu2020robotic} or electrocution~\cite{ascard200710}, and more recently laser-based~\cite{xiong2017development} implement.
These offer different benefits and restrictions.
Vahdanjoo~\etal~\cite{vahdanjoo2023operational} presented a comprehensive analysis of the operational, economic, and environmental effects of the Robotti-Intelli platform in seeding and weeding management in arable farms.

In~\cite{zhou2021design} Zhou~\etal\ only focused on the design and preliminary evaluation of a targeted spray platform. 
All components were effectively evaluated, especially regarding the response time and target spray accuracy.
They conducted only an indoor experimental setup showing that the developed system can reduce $46.8\%$ usage of chemicals compared to the uniform spray method.
Similarly, a mechanical weeding tool was deployed in~\cite{Chang21_mechanical_tools} where they operated on a self-built robotic platform on a single-row early-grown vegetable field.
This hoeing mechanism brought out the roots of weeds but due to limited work space and length of operation, they are only deployable on specific crops with certain growth stages.
The effectiveness of weed management tools varies based on field conditions and plant types. 
While recent weeding robots surpass traditional methods in precision, still there is a large space for improvement. 
As this technology evolves, it exemplifies how innovation can transform traditional farming into smarter, eco-friendly agriculture.
Our method with \bbot\ addresses this by using advanced sensors and AI to precisely target weeds, optimizing efficiency and sustainability in agricultural practices. 


\section{Selective Precise Intervention}
\label{selectivePreciseIntervenstion}
\input{contribs/selectivePreciseIntervenstion.tex}


\section{Experimental Setup}
\label{sec:exp_steup}
\input{contribs/expSteup}

\section{Experimental Evaluations}
\label{sec:exp}
\input{contribs/experiments}

\section{Conclusion}
\label{sec:conc}
In this paper, we introduced advancements in weed management, enhancing the capabilities of \bbot~for conducting precise bio-diversity-aware weeding of sugar-beet fields. 
We introduce an advanced planning method that employs a rolling-view technique, leading to an average absolute performance enhancement of $3.4\%$. 
Furthermore, our experiments in real-fields with real-robot demonstrated that \bbot~ weeding strategy could be effective only by having a loss of $11.66\%$ due to planning or intervention constraints.
We present, for the first time, a concept of bio-diversity-aware weed management and assess its practicality in real-world scenarios.
Our approach not only improves the precision and effectiveness of agricultural robots but also emphasizes the necessity of incorporating ecological considerations into crop management.
In conclusion, our research indicates that performance variations in weeding technologies are often due to dynamic real-field conditions, such as unexpected weather shifts impacting weed detection and treatment accuracy. 
Future work will aim to enhance these systems' adaptability to such environmental factors, focusing on developing more robust and responsive vision systems to ensure consistent and effective weed management in diverse agricultural settings.


\section*{Acknowledgements}

This work was funded by the Deutsche Forschungsgemeinschaft (DFG, German Research Foundation) under Germany’s Excellence Strategy - EXC 2070 – 390732324.

\bibliography{references}
\bibliographystyle{IEEEtran}

\end{document}

%% file: contribs/selectivePreciseIntervenstion.tex
\begin{figure}[!t]
	\centering
        \vspace{2mm}
	\includegraphics[width=1.0\linewidth]{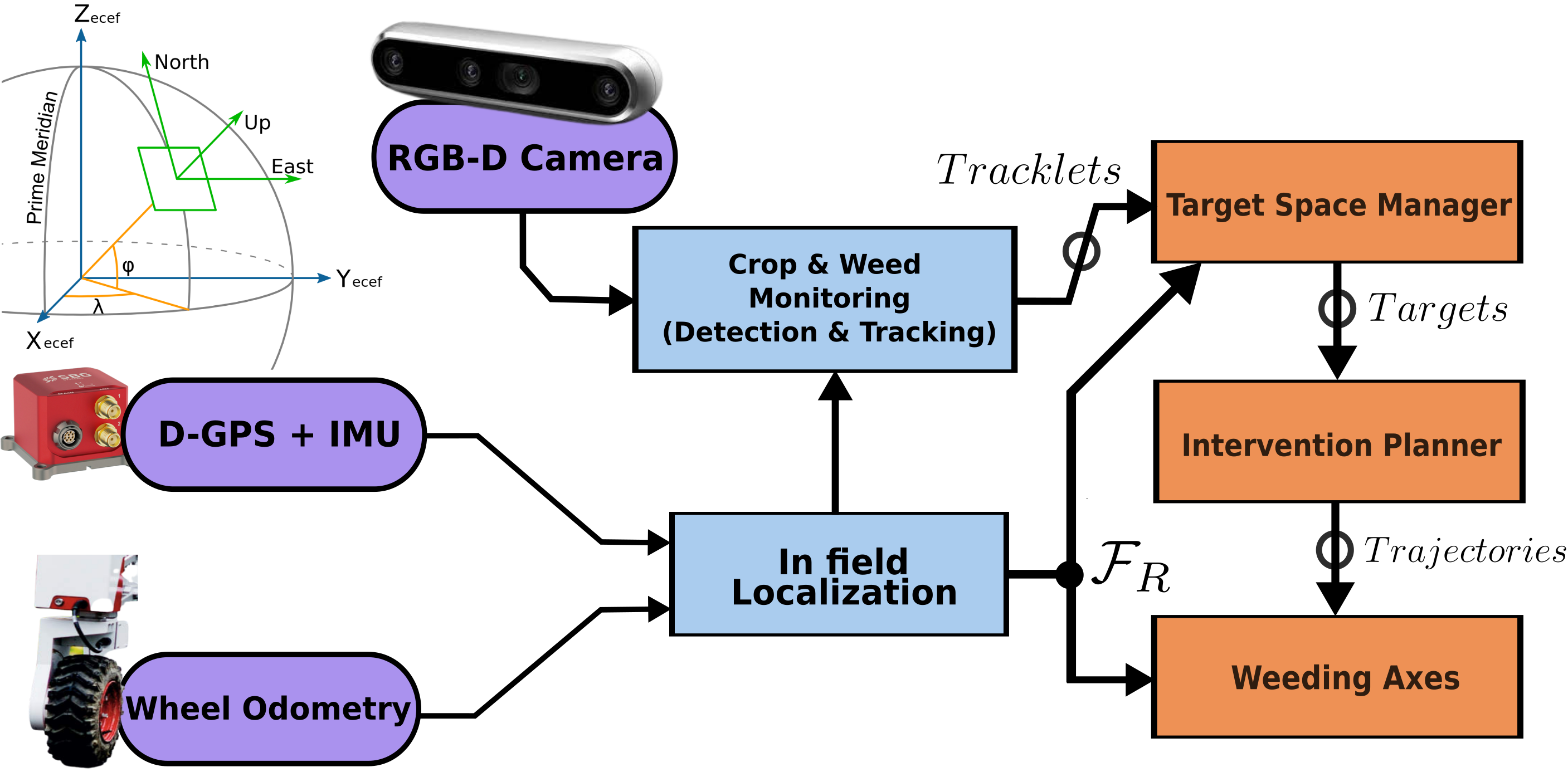}
	\caption{The software architecture, including sensors (purple), vision perception, localization of robot base frame $\mathcal{F}_R$ (light blue), intervention planning, intervention planner, and weeding axes controllers (orange).}
	\label{fig:inferenceModel}
 	\vspace{-6mm}
\end{figure}
\bbot~\cite{ahmadi2022bonnbot} is a versatile agricultural platform designed at the University of Bonn.
It is equipped with a novel weeding tool design enabling high-precision plant-level field interventions. 
In our previous work~\cite{ahmadi2022bonnbot} we elaborated on the conceptual design of the weeding tool and its associated software components.
\figref{fig:inferenceModel} shows the software architecture of~\bbot, where, different sensors available in the robot are supplied to the localization and crop-weed monitoring modules to enable precision interventions through the weeding tool.
The controller system uses ROS (Robot Operating System) to standardize node communications.
The weeding tool features four independently controlled high-resolution linear actuators positioned $0.6m$ above the ground, creating a working area of $1.39m\times0.36m$. 
Currently, they are equipped with spot-spray nozzles (with a spray footprint of 5cm on the ground) which enable them to be used independently to engage with weeds.
For more details about \bbot\ platform, we refer the reader to~\cite{ahmadi2022bonnbot}.

The most critical challenge in this design is to efficiently plan paths for intervention heads, such that we maximize the number of well-managed weeds in a weeding scenario.
A typical weeding scenario is composed of the following stages:
\textbf{\RNum{1}} Weeds get detected, classified, and tracked within the viewable area of the down-facing detection camera in front of the robot.
The monitoring system which runs the Mask-RCNN network for instance-based semantic segmentation and classification is used to estimate necessary phenotypic information about the plants.
Finally, the plants are tracked (tracklets) through the viewable area, more details can be found in~\cite{ahmadi2022bonnbot}.
\textbf{\RNum{2}} The tracklets, which are identified as valid plants for management, are monitored beneath the robot using its localization system. 
These tracklets are then relayed to a target-space manager, who efficiently allocates them among the various weeding axes for optimal handling.
\textbf{\RNum{3}} Each intervention planner plans an optimal path considering phenotypic factors like weed type, size, harmfulness, required action time, and bio-diversity considerations.
\textbf{\RNum{4}} The computed paths are transmitted to the relevant axis drivers and nozzle controllers to carry out corresponding actions on the plants. 
\subsection{Crop-Weed Monitoring Pipeline}
\label{subsec:envPreception}
\input{contribs/envPreception}

\subsection{Rolling-view Observation Model}
\label{subsec:observationModel}
\input{contribs/obsModel}

\subsection{Target-Space Management}
\label{subsec:targetSpaceManagement}
\input{contribs/targetSpaceManagement}

\subsection{Bio-Diversity-Aware Plant-level Treatment}
\label{subsec:bioDivAwareTreatment}
\input{contribs/bioDivAwareTreatment}

\subsection{Intervention Heads Route Planning}
\label{subsec:routePlanning}
\input{contribs/routePlanning}

%% file: contribs/envPreception.tex
The vision system on \bbot\ uses instance-based semantic segmentation based on the Mask-RCNN~\cite{he2017mask} architecture.
We use a variant of Mask-RCNN, initially proposed by us in~\cite{halstead2018fruit}, which provides a super-class classification (plant versus background), a sub-class classification (crop and weed species), bounding box location, and pixel-wise segmentation for each object in the scene.
Providing these classification results and pixel-wise labels allows for the vision system to drive selective intervention tasks so that weeds are only sprayed if they meet certain requirements.

To train Mask-RCNN with sub-class labels we use our novel dataset SB21 and two previously released crop and weed datasets: SB20~\cite{ahmadi2021virtual} and WeedAI~\cite{halstead2024}.
In each dataset, several variations occur naturally, including, weed species, crop and weed densities, illumination variation, and shadowing caused by the platform.
These variations make the model able to generalize better to new scenes. To further generalize our approach we also include strong data augmentation.

We employ multiple data augmentation techniques such as luminance variation and blur, additionally, we propose an ``illumination box`` approach which simulates self-shadowing effects.
For all of our augmentation techniques, we set a probability of 0.5 that they will be used on an individual basis.
To augment the image luminance we randomly vary the L channel of the Lab color space between $\pm{15}$.
RGB jitter can also be selected with a random color channel variation of $5\%$ and to simulate noise in the scene we also randomly employ  Gaussian blur with a $3\times{3}$ kernel.
Finally, as we see shadowing caused by the robot in the WeedAI dataset and to ensure consistency with the other two datasets we randomly insert an ``illumination box''.
This ``illumination box'' replicates the shadowing effect by randomly inserting a luminance increase (in the LAB color space) over the area of a random rectangular box.
As the natural shadowing does not produce clean edges nor perfectly horizontal boxes we also blur the edges of the rectangle using Gaussian blur with a kernel size of $3\times{3}$ and a random rotation of the box to $\pm{10^o}$.
These augmentation techniques aim to artificially extend our datasets while also ensuring better generalization.

One of the tasks we perform here is a weeding intervention based on informed decisions.
These informed decisions come from the type of plant being segmented (crop or weed) and the size of the plant (growth stage).
To achieve this level of information we utilize the classification scores and the masks generated by Mask-RCNN, and the registered depth information captured from our RGB-D sensor.
This enables us to estimate the growth stage of a plant by converting the depth to a per-pixel area and accumulating over the pixel-wise instance-based semantic segmentation mask, in a similar method to that described in~\cite{halstead2021crop}.

The final step in our vision system is the tracking component, which is based on our prior work of crop-agnostic monitoring by Halstead \etal~\cite{halstead2021crop}.
We exploit the depth and odometry information to reproject our pixel-wise segmentation to perform better tracklet matching.
This reprojection can better align new masks with existing tracklets.
We also employed the dynamic radius tracking approach since it improves the tracking of small plants in our complex scenes.

%% file: contribs/obsModel.tex
\begin{figure}[!t]
	\centering
        \vspace{2mm}
	\includegraphics[width=0.95\linewidth]{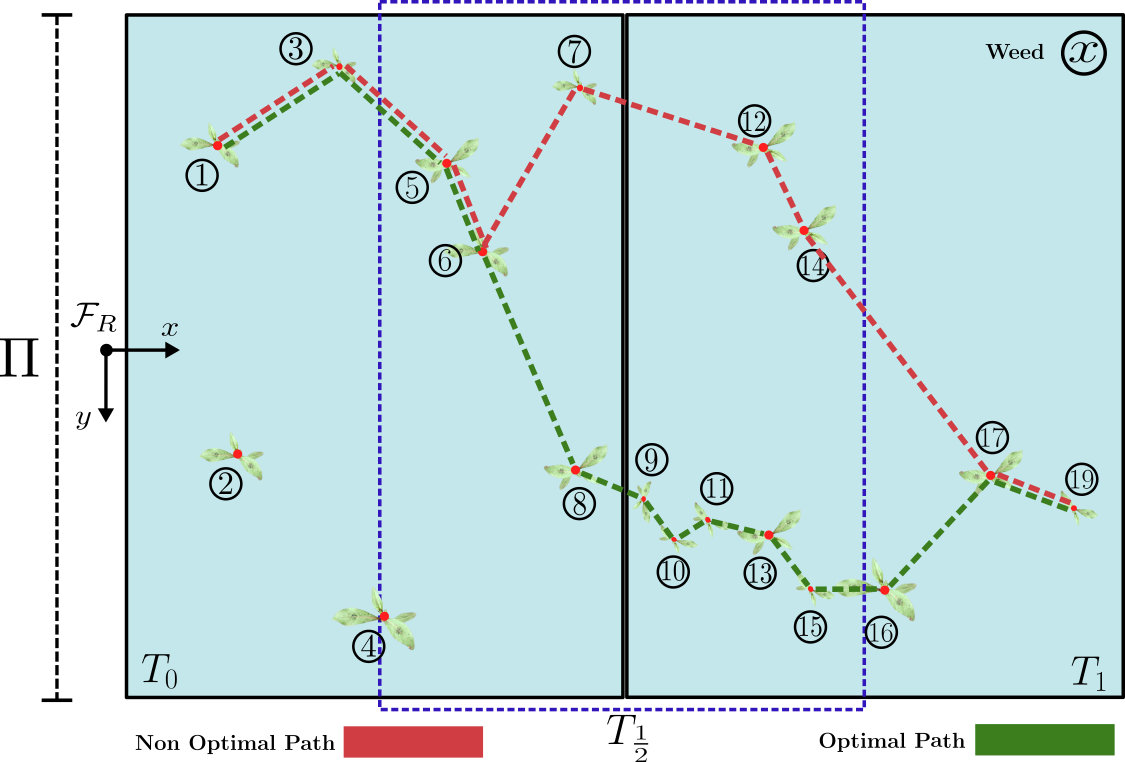}
	\caption{Segment-view vs Rolling-view Planning; two separate segments $T_0$ and $T_1$ with different weeding distributions are shown. An intermediate $T_i$ is substantially helping the planner to optimize the planned route of the weeding axis (green) w.r.t the baseline (red).}
	\vspace{-6mm}
	\label{fig:rollingVsDiscrete}
\end{figure}

In our previous work~\cite{ahmadi2022bonnbot} we developed a segment-view observation model.
This segment-view-based model allows us to focus on an individual field segment, enabling us to cope with their distinct weed distributions. 
We derive the distance between the weeding equipment and individual weeds assuming a weed distribution yielded by a Poisson process with an arrival rate of $\eta =\lambda \times \Pi$, where $\Pi$ denotes the weeding width illustrated in~\figref{fig:rollingVsDiscrete}.
To understand the spacing between the weeds ($\delta_x$) we use the robot's motion along the $x$-axis within its frame of reference ($\mathcal{F}_R$) depicted in~\figref{fig:rollingVsDiscrete}.
While this method has proven its worth, we believe that it is less adaptable to the changing weed spatial distribution patterns and thus harms overall performance. 
To overcome this we propose a rolling-view observation model.
This model integrates discrete camera observations into a comprehensive model of the entire field.
This new approach integrates discrete camera observations into a comprehensive model of the entire field. 
Expanding the planning scope beyond single segments, allowing for more effective weeding strategies over a larger area.
A key advantage of this approach is illustrated in Figure~\ref{fig:rollingVsDiscrete} which highlights that the previous segment-based approach would treat two views independently whereas the rolling window approach updates the planned intervention across these two views iteratively. 
This enables the planning to be optimized more flexibly without increasing the planning window size.
By simply incorporating only one intermediate frame~$T_\frac{1}{2}$, we achieve a notable enhancement in the weeding trajectory and overall performance.
We have implemented this algorithm using multi-threading and dynamic programming techniques that enable the rolling-view planning to be at a frequency $> 500$Hz on CPU (Intel® Core™ i7-12700K).
The rolling window planner uses the most recent information. This means we take the prior plan and update it only if there are new plants that need to be treated.
This rolling-view model not only refines the weeding process and inter-image tracking but also compensates for the vision system's shortcomings, like missed detections or incorrect classifications. 
The accumulation of multiple detections over the same area considerably increases the accuracy of our predictions.
Furthermore, by analyzing continuous field segments, we can predict more precise weeding paths, leading to improved overall efficiency in weed management.

%% file: contribs/targetSpaceManagement.tex
When a robot encounters multiple targets $\mathcal{N}$ beneath it, the primary objective is to engage each target using one of its intervention heads $\mathcal{H}$ while the robot is in motion. 
Hence, before these targets enter the workspace of the weeding tool, it's crucial to plan the motion for each intervention head. 
Considering sets of weeds in any given workspace motivates a multi-query approach to the problem.
We consider weeds to be presented as a set that motivates us to assign them to the $\mathcal{H}$ intervention heads.

Hence, we use a high-level planner to distribute targets between different intervention heads while ensuring that all the targets will be visited at least once and the workload is properly balanced between all weeding axes. 
In~\cite{ahmadi2022bonnbot}, we propose the Distance-based Target Assignment, Static Work-space Division-based Target Assignment, and Dynamic Work-space Division-based Target Assignment. 
In this paper, we will only use the static work-space division-based target assignment approach which based on our real field experiments offers stable and reliable performance.

%% file: contribs/bioDivAwareTreatment.tex
\begin{figure}[!t]
        \vspace{2mm}
	\centering
	\includegraphics[width=0.75\linewidth]{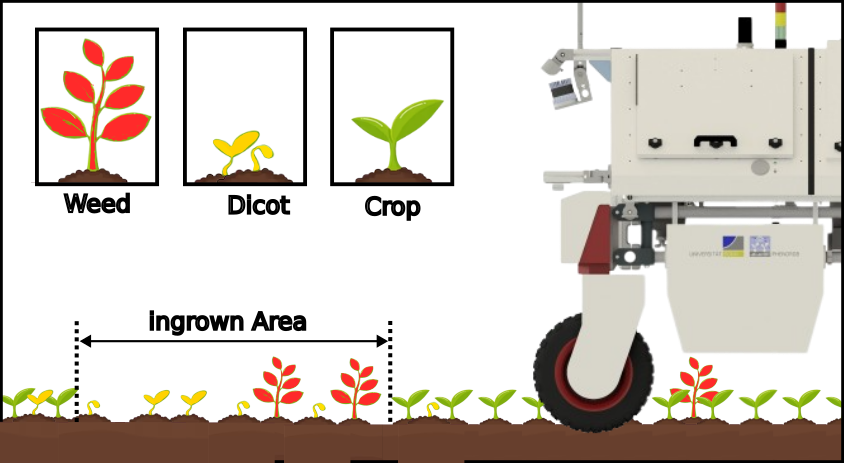}
	\caption{Biodiversity in Focus: A field area with low crop density, susceptible to invasive weeds, and the potential role of Dicot plants in controlling weed growth.}
\label{fig:bioDiversityScheme}
    \vspace{-5mm}
\end{figure}

Weed harmfulness is influenced by aggressiveness, competition for water, and leaf size, affecting nearby crops. 
Species like barnyardgrass, pigweed, and lambsquarters can significantly outcompete crops for light, water, and nutrients, reducing yield and quality~\cite{zimdahl2007weed}. 
In moisture-limited environments, this competition becomes more critical. Understanding these factors is essential for effective weed management, recognizing that not all weeds are detrimental; some can enhance soil fertility and moisture retention.
An example scenario is depicted in~\figref{fig:bioDiversityScheme} where dicots are considered beneficial.

Therefore, robots must recognize and adapt to bio-diversity when managing these varied weed threats. 
In this case, a generalized approach to weed management won't suffice due to the diverse competitive behaviors of different weeds. 
We propose a novel weeding method enabling \bbot\ to consider bio-diversity factors by differentiating between weed species and their phenotypic characteristics, assessing their threat levels based on competitiveness factors, and acting accordingly. 
This ensures not only the well-being and productivity of crops but also resource optimization concerning weeding actions (i.e. spraying). 

Relying upon the phenotypic information, we aim to facilitate our robots' understanding of different weeding scenarios in the real field.
Using the monitoring technique introduced in~\secref{subsec:envPreception} we determine the type, growth stage, and distance of weeds in a cropping area. 
We introduce the 'harmfulness' factor~$\kappa$, illustrated in~\eqref{eq:harmfulnessCoef}, for each weed type, ranking them based on their threat to the nearby crops. 
It is important to acknowledge that currently, there is no definitive method for assessing plant-wise harmfulness effects in actual fields on different crops. 
However, in this work, we aim to build upon existing heuristics to facilitate incorporating this information into future weeding strategies.
Hence, we derived this factor from phenotypic data, which helps our robots discern various weeding scenarios. 
It ranks weeds based on their threat to crops, using a methodology inspired by Dentika~\etal~\cite{dentika2021weeds}, who proposed a similar concept for pathogen-host and disease risk imposed by weeds on the crops, and was derived from comprehensive on-site experiments and is formulated below:
\begin{equation}
    \kappa(w, p) = \dfrac{\alpha_w \; \beta_w \; }{\alpha_p \times \Delta(w, p)}
    \label{eq:harmfulnessCoef} 
\end{equation}
In~\eqref{eq:harmfulnessCoef}, the harmfulness effect $\kappa(w, p)$ of weed $w$ on plant $p$ is calculated where, $\alpha_w$ and $\alpha_p$ denote the size of weed $w$ and crop $p$ in $mm^2$, respectively.
This is reflecting the competitive relationship between a weed and its adjacent crop, based on their sizes.
Also, $\beta_w$ is the harmfulness factor of each specific weed category, noting that certain weeds should be eliminated from the field regardless of their size or specific location.
Furthermore, the $\Delta(w, p)$ is the Euclidean distance between crop $p$ to weed $w$. 
This acknowledges that weeds situated far from crops may not pose a significant threat, thus requiring no intervention.
This approach promotes eco-friendly interventions by prioritizing the most harmful weeds, thereby enhancing biodiversity and resource conservation.

%% file: contribs/routePlanning.tex
We need to generate $\mathcal{H}$ independent and efficient routes to guide the intervention heads, taking into account assigned targets, the intervention head's position, the robot's linear speed, and the speed and acceleration limits of its linear axes.

To do this, the intervention controller embeds all the details about each plant in a uni-directional graph, including its type, segmentation, estimated size, boundaries, plant center, and corresponding bio-diversity characteristics. 
In this graph, each node represents a detected plant and every edge scores the motion feasibility between two nodes.
As introduced in~\cite{ahmadi2022bonnbot}, the probability of visiting the $j$-th weed with $i$-th weeding nozzle is computed as follows,
\begin{equation}
    \label{eq:weedingProb}
    \textit{P}_{ij} = \textit{P}\left( \dfrac{\gamma}{\vartheta}  <   \dfrac{\delta^{x}_{ij}}{\delta^{y}_{ij}} \right),
\end{equation}
where $\vartheta$ and $\gamma$ denote the maximum velocity of linear axes and maximum linear velocity of the robot, $\delta^{x}$ and $\delta^{y}$ denote the relative distance between the $i$-th nozzle to the $j$-th weed.
In this scheme, we define each edge to consist of a feasibility score~$\Gamma_{jk}$ using the logistic function~\eqref{eq:logisticFunc} of motion between nodes (weeds)~$j$ and~$k$.
\begin{equation}
    \Gamma_{jk} = \frac{1}{1 + e^{-\omega(S_{jk})}} ,
    \label{eq:logisticFunc} 
\end{equation}
where $S_{jk}$ is the favorability score in seconds
\begin{equation}
    S_{jk}=\dfrac{\delta^{x}_{jk}}{\gamma} - \dfrac{\delta^{y}_{jk}}{\vartheta},
    \label{eq:favorability} 
\end{equation}
and the weighting parameter $\omega$ adjusts how quickly the favorability score makes the~$\Gamma_{jk}$ change from the boundary score (0.5) to being very likely (1.0).
The boundary score occurs when $S_{jk}=0$ and represents when there is just enough time for the tool to transition from node $j$ to $k$.

In a greedy algorithm, every potential route is calculated by permuting all nodes in the graph, considering the edge directions in the node graph. 
The best route is the one that visits a high number of nodes with high feasibility.
The best performed method utilized in~\cite{ahmadi2022bonnbot} was $n$OTSP which is a modification of the conventional traveling salesman problem. 
The planning strategies generate a set of $m$ possible trajectories~$\Vec{\mathbf{T}}=[\Vec{T}_0,\dots,\Vec{T}_{m-1}]$, where each~$\Vec{T}_t$ represents an organized list of weed locations in the trajectory, consisting of $q_{t}$ elements in form of~$\Vec{T}_t=[w_0,\dots,w_{q_{t}-1}]$.
Using the following criteria we calculate the success criterion $\Vec{\mathcal{C}}=[\mathcal{C}_0,\dots,\mathcal{C}_{m-1}]$ for each trajectory in $\Vec{\mathbf{T}}$.
\begin{equation}
    \label{eq:trajectoryProb} 
    \Vec{\mathcal{C}}(\Vec{\mathbf{T}}, \rho) = \left\{\begin{array}{cc}
	    \dfrac{1}{q_t} \sum\limits_{r=0}^{q_{t}-1} \Gamma(n_{r}, n_{r+1}), & for \ all  \ \Gamma \ge \rho \\
	     0 & otherwise
	\end{array}\right.
\end{equation}
where $\rho = 0.6$ is a cutoff threshold applied to each feasibility score ensuring the planner only considers trajectories with all reachable targets.
To incorporate the harmfulness factor, we use the $\Vec{\mathbf{T}}$ where $\mathcal{C}_t > 0$, and pick the trajectory with maximum total harmfulness score $\mathcal{K}$ using,
\begin{equation}
    \label{eq:trajectoryCost} 
    \argmax_{t} \mathcal{K}(\Vec{\mathbf{T}}); \;\; \mathrm{where} \;\; \mathcal{K}(\Vec{T}_{t})=\sum_{r=0}^{q_{t}-1} \kappa \left(\Vec{T}_{t}(r) \right)
\end{equation}
and $\kappa_r$ is the harmfulness factor, \eqref{eq:harmfulnessCoef}, of the $r$-th node in $\Vec{T}_t$.
This leads to the best trajectory being the one that has the largest number of reachable targets and high-priority weeds.

%% file: contribs/expSteup.tex
Similar to our prior work~\cite{ahmadi2022bonnbot}, we consider that the robot moves with constant speed $\gamma=0.5m/s$ along a crop row with weed density of $\lambda$ weeds/$m^2$ and we set the velocity of the linear actuators to $\vartheta=5m/s$.
All experiments are conducted using 4 linear axes.

\subsection{Real-Field Models}
The field models are sourced from the crop monitoring method detailed~\secref{subsec:envPreception} and compiled into a representative row format. 
We generate models from three real-world fields captured on either sugar-beet or corn crops at CKA across two years.
This creates complex scenes to evaluate as there are large differences in the weed densities across the datasets.
These models contain four distinct weed densities: low (CN20), moderate (SB20-S2), high (SB20-S1, SB21-S1), and very high (SB21-S2).
To evaluate the performance of conducted operation both in simulation and real-fields we use the number of untreated weeds and from now on we call it loss (in percentage).
Furthermore, the traveled distance (in meters) of the linear axis could provide us with useful information about the load balance and effectiveness of planning.

\subsection{SB21 Dataset}
\label{subsec:dataset}
We utilized BonnBot-I to create a series of distinct datasets tailored for precision agricultural applications. 
In our prior works, the datasets were acquired at sugar-beet (SB20)~\cite{ahmadi2021virtual} and corn (CN20)~\cite{ahmadi2022bonnbot} fields within the University of Bonn's Klein-Altendorf campus (CKA). Both datasets featured sparsely annotated video sequences with no overlap between annotated frames.
In this work, we introduce a new sugar-beet dataset (SB21) captured at CKA during 2021. 

\begin{figure}[!t]
        \vspace{2mm}
        \centering
	\begin{tabular}{cc}
        \includegraphics[width=0.48\linewidth]{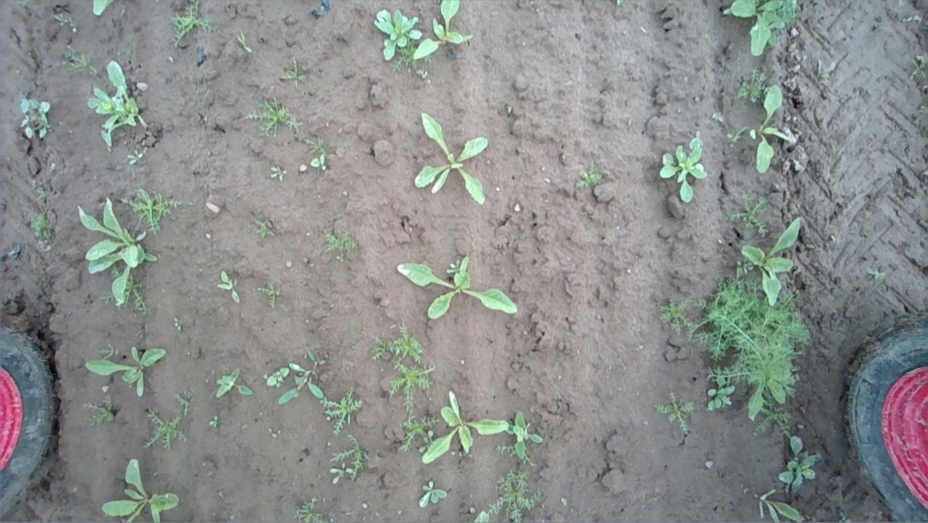}
        \includegraphics[width=0.48\linewidth]{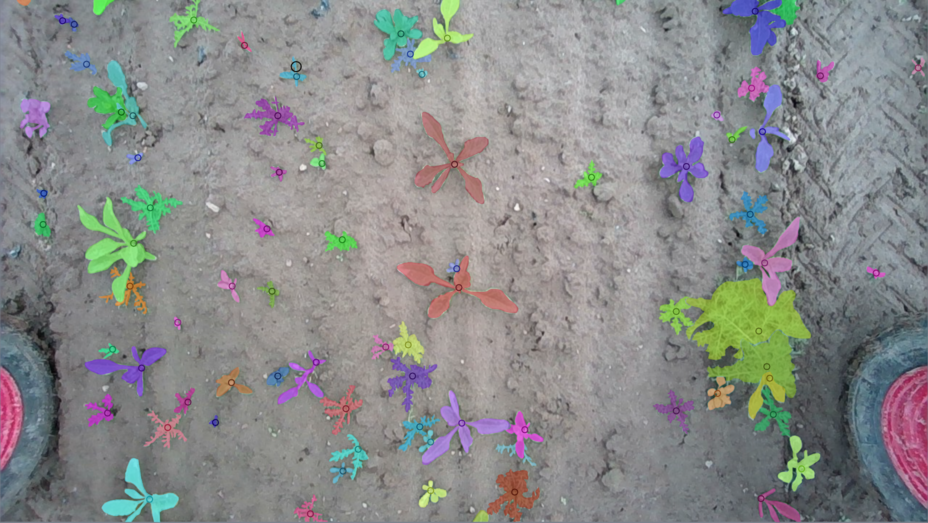}
    \end{tabular}
    \caption{Example image of dataset \textit{SB21} (left) and multi-class annotations representing different types of crops and weeds using different colors (right).}
	\label{fig:datasetSample}
	\vspace{-5mm}
\end{figure}

SB21 is comprised of 84 RGB-D frames of crops and $12$ distinct weed categories, totaling $942$ instances of sugar beet and 4989 instances of weeds. 
The images are divided into train, validation, and evaluation sets with respectively 56, 14, and 14 images.
Annotations include pixel-wise segmentation on an instance basis, bounding boxes, and stem locations for each instance. 
Similar to SB20 and CN20, the images are augmented with robot pose and velocity information from \bbot. 
The data structures proposed here enable the creation of real temporal sequences, where only the final frame in the sequence is labeled, as detailed in~\cite{ahmadi2021virtual}, thereby reestablishing temporal relationships between annotated and non-annotated frames.
\figref{fig:datasetSample} provides an example of an annotated image from the \textit{SB21} dataset.

%% file: contribs/experiments.tex
In this section, we present the outcomes of various experiments conducted both in simulated and real-world settings.

\subsection{Segment-view observations vs Rolling-view observations} 
\begin{table}[b!]
        \vspace{-3mm}
	\centering
	\caption{Weeding performance of \bbot\ with two observation models Segment-view and Rolling-view denoting weeding loss ($\%$) and traveled distance ($m$) of the linear axis.}
	\resizebox{\columnwidth}{!}{%
	\begin{tabular}{l | cc|cc|cc}
	\toprule
	 & \multicolumn{2}{c|}{\textbf{Weed-density}}& \multicolumn{2}{c|}{\textbf{Segment-view}}  & \multicolumn{2}{c}{\textbf{Rolling-view}} \\
	 & - & Avg/$m^2$ & ($\%$) & ($m$) & ($\%$) & ($m$) \\ \hline	 
	 \midrule
    CN20        & low       & 3.1  & \textbf{0.0} & 2.7$\pm$0.2    & \textbf{0.0}  & 2.0$\pm$1.6 \\
    \midrule
    SB20-S1     & moderate  & 8.2  & \textbf{0.0} & 1.4$\pm${0.2}  & \textbf{0.0}  & 1.9$\pm$0.9 \\
    SB20-S2     & high      & 15.4 & 11.9         & 10.1$\pm${0.9} & \textbf{6.4}  & 4.1$\pm$2.1 \\
    \midrule
    SB21-S1     & high      & 22.3 & 17.2         & 6.1$\pm$5.3    & \textbf{14.1} & 2.8$\pm$2.6 \\
    SB21-S2     & very high & 81.2 & 38.2         & 13.4$\pm$8.2   & \textbf{36.5} & 5.2$\pm$3.1 \\
    \bottomrule
    \end{tabular}}
    \label{tab:realPlanning}
\end{table}
First, we illustrate the efficiency of the weeding system on real-field models, showcasing the $n$OTSP approach in two distinct observation modes, Segment-view (baseline) and our novel Rolling-view observations. 
Our evaluations are applied to the test rows within the three datasets (CN20, SB20, and SB21).
\tabref{tab:realPlanning} summarizes the difference in planning performance and traveled distance of weeding axes between the two observation models (Segment-view (baseline) and Rolling-view) deployed on the real-field models.
Below we briefly describe the results in~\tabref{tab:realPlanning} in terms of weed density, from low to very high.

\textbf{Low weed density}: both methods demonstrate comparable performance in areas with lower weed density achieving zero loss on CN20 and leaving no untreated weeds in the fields. 
The Rolling-view model, however, required less travel distance compared to the Segment-view. 
However, the variance in the traveled distance of the weeding axes seems to be lower when using Segment-view planning. This suggests a more balanced distribution of targets across the different implements.
\textbf{Moderate weed density:} Similar to low density, both models achieved $0.0\%$ loss, with no noticeable difference in weeding efficiency. 
However, the Rolling-view model had a slightly higher travel distance ($1.9\pm0.9m$) than the Segment-view ($1.4\pm0.2m$).
\textbf{High and very high weed density:} In this scenario, the Rolling-view model outperformed the Segment-view model. 
The Rolling-view achieved only a $6.4\%$ loss compared to the Segment-view's $11.9\%$ loss. 
Additionally, the Rolling-view required a substantially lower travel distance than the Segment-view, showing the efficiency of the Rolling-view method in more complex situations.
Overall, the Rolling-view planning method surpasses the baseline with an average absolute improvement of $3.5\%$ over all the field models with a maximum and minimum absolute improvement of $5.5\%$ on SB20-S1 and $1.7\%$ on SB21-S2, respectively.

\subsection{Bio-diversity aware weeding operation} 
The bio-diversity-aware system should accurately differentiate between crops and weeds.
This includes detecting beneficial dicots and only intervening where it is essential for crop health while preserving and promoting biodiversity.

This weed classification approach is particularly crucial in challenging scenarios.
These scenarios can include high weed density, restricted robot velocity, restricted linear axis speed, or herbicide usage limitations.
Hence, the system must manage certain losses while achieving optimal performance. 
To test our system's resilience and capability under extreme conditions, we devised an experiment where the robot operates at a speed two times the normal $\gamma=1.0 m/s$. 

In this analysis, we utilized the field models from our previous experiment but with two modifications.
First, we designate two priority levels where a low-priority for weeding is given to potentially beneficial dicots and a high-priority is given to all other weeds.
To represent the priority levels we give a weight of 0.1 and 1.0 for low- and high-priority respectively.
Second, the occurrence of high-priority weeds was set to be one-tenth of that of low-priority. 
Consequently, this setup required the bio-diversity-aware system to adjust its weeding trajectories to prioritize the less frequent, high-priority weeds, even if it meant potentially overlooking some of the low-priority weeds.
In the following, we compare the bio-diversity-aware (Bio-Div.) approach with a baseline (not-bio-diversity-aware) method under extreme conditions on the same real-field models used in the previous experiment. 
\begin{figure*}[!t]
    \centering
    \vspace{2mm}
    \efbox{\includegraphics[width=0.65\linewidth]{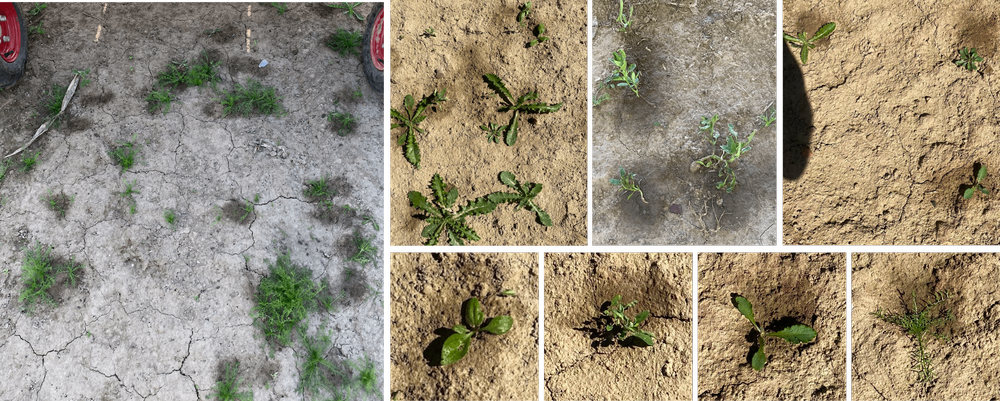}}
    \caption{Real-field intervention examples; Different spray footprints of BonnBot-I weeding operation in campus CKA of the University of Bonn, with different weed densities and various weed types.}
    \label{fig:real_field_intervention_footprints}
    \vspace{-5mm}
\end{figure*}
The weeding loss for the Bio-Div. the approach is very similar to the baseline.
It leaves no untreated weed in the fields, indicating its comparable effectiveness in scenarios with low to moderate weed densities.
However, in cases of very high weed densities (SB21-S1, SB21-S2), there is a minor ($<2\%$) increase in weeding loss with the Bio-Div. approach.
This suggests a small trade-off in performance at extremely high weed densities. 
The average traveled distance of axes using both observation methods is generally similar or marginally higher ($<0.2m$ on average) for the Bio-Div. approach across all field models, reflecting the method's thoroughness in targeting specific weed priorities and achieving bio-diversity considerations.
Additionally, to enhance our evaluation of the operation's specifics, we detailed the treatment percentage for each weed type individually and their results are presented in~\tabref{tab:weedCatsHits}.
\begin{table}[b!]
    \vspace{-4mm
    }
	\centering
	\caption{Weeding performances (treatment percentage) of \bbot\ with baseline weeding and Bio-diversity-aware schemes on real-field models.}
    \resizebox{\columnwidth}{!}{%
    \begin{tabular}{l | cc|cc}
    \toprule
        &
        \multicolumn{2}{c|}{\textbf{Weed 1 }(Low Priority)}& 
        \multicolumn{2}{c}{\textbf{Weed 2 }(High Priority)} \\
        Field Model & (not-Bio-Div.) & (Bio-Div.) & (not-Bio-Div.) & (Bio-Div.) \\	 
        & ($\%$) & ($\%$) & ($\%$) & ($\%$) \\ \hline	 
    \midrule
    CN20         & 73.3 & 73.3 & 56.6 & 56.6 \\
    \midrule
    SB20-S1      & \textbf{47.5} & 44.3 & 56.0 & \textbf{60.2} \\
    SB20-S2      & \textbf{42.5} & 40.5 & 53.5 & \textbf{63.5} \\ 
    \midrule
    SB21-S1      & \textbf{18.0} & 15.2 & 19.1 & \textbf{38.8} \\
    SB21-S2      & \textbf{7.0}  & 5.1  & 7.0 & \textbf{17.4} \\
    \bottomrule
    \end{tabular}}
    \label{tab:weedCatsHits}
    
\end{table}

When considering the low-priority weeds, the Bio-Div. method achieves slightly reduced performance when compared to the baseline.
On the other hand, this reduced performance could be interpreted as a positive point, where keeping low-priority weeds in the field could benefit the ecosystem and align with bio-diversity purposes.
However, the benefit of this approach is evident in the ability of the Bio-Div. approach to achieve stronger performance when targeting higher-threat weeds.
Our approach is able to improve performance on all but the lowest weed distribution (CN20) where performance is identical.
This illustrates the Bio-Div. approach's effectiveness in prioritizing and managing high-priority weeds.



\subsection{Real-World Intervention Performance} 
In~\cite{ahmadi2022bonnbot} we only evaluated the weeding performance on recorded data and considered non-overlapping segments of observations for running the experiments.
To further evaluate the whole system's performance, we deployed BonnBot-I in a series of unseen fields with different weed distributions, illumination conditions, and cultivars. 
The experiments were conducted on various days, under a mix of weather conditions including partial cloudiness and sunshine, and on different types of soil ranging from solid and compact to relatively muddy. 
Additionally, the plants tested were at various growth stages, from the two-leaf stage up to the eight-leaf stage.
The field experiments with the real robot were conducted at the CKA, covering nearly 100 square meters of sugar-beet field. 
This evaluation outlines the most important metrics associated with the operation of the robot.
These metrics include the number of missed weeds, the number of partially treated weeds, and the number of accurately treated weeds.
We quantitatively evaluated our performance on two different weeding scenarios: crop-weed; and weed-only.

\begin{figure}[!t]
    \centering
    \vspace{2mm}
    \includegraphics[width=1.0\linewidth]{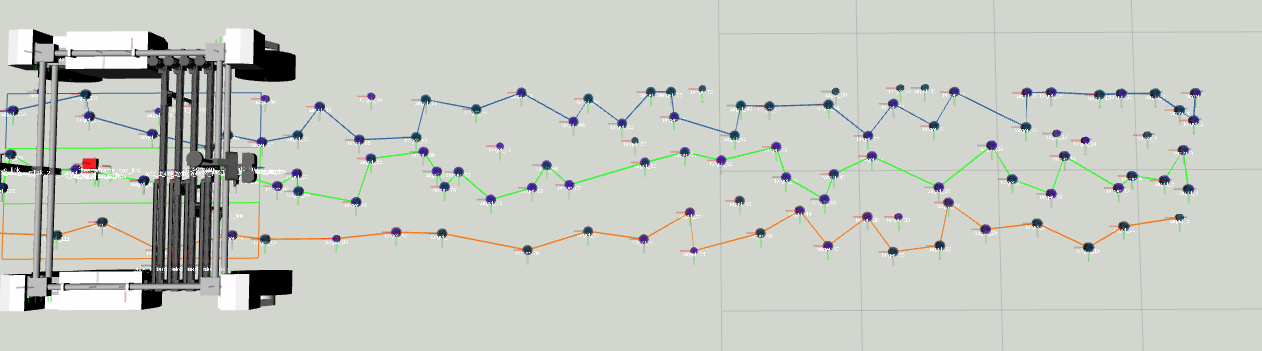}
    \caption{BonnBot-I Simulation; An example field with weed density $\lambda\approxeq12$ with planned trajectories of three linear axes depicted in different colors (blue, green, and orange).
    All dots that lie along one of the lines are weeds that will be treated, while all the dots that do not lie upon a line are not able to be treated.}
    \label{fig:simulation_weeding_trajectories}
    \vspace{-5mm}
\end{figure}

During all evaluations, we used the Rolling-view observation model enabling more accurate planning in real-world scenarios.
An example of this planning approach is depicted in~\figref{fig:simulation_weeding_trajectories} where only three linear axes are utilized (indicated by the three colors).
The weed-only field model was used to evaluate our intervention approach independently of vision system failures (i.e. classifying a weed as crop or vice versa).
Our quantitative analysis is based on visually interpreting the footage captured on a camera mounted at the rear of \bbot.
\figref{fig:real_field_intervention_footprints} demonstrates the footprint of weeding interventions in the real fields.

The trial fields contained regions with low ($<5$) to high weed ($>10$) densities, therefore offering challenging scenarios for testing \bbot\ weeding capabilities.
In total our two fields contain $1038$ weeds, the sugar-beets in the crop-weed fields had plants in the four- to six-leaf stage (approximately three to six weeks old).
\tabref{tab:realPerf} summarizes the performance of \bbot\ in both weed-only and crop-weed regions.
Initial evaluations in the crop-weed region showed, that \bbot\ has treated $23$ crops falsely which counts as a total of $4\%$ of visited crops in the field.
Furthermore, of the total weeds in the fields, we accurately detected $886$.
Of these detected weeds \bbot\ did not perform an intervention on $121$ of them.

\begin{table}[b!]
    \vspace{-4mm
    }
	\centering
	\caption{Real-world weeding performance of \bbot\ in weed-only and crop-weed regions.}
    \resizebox{\columnwidth}{!}{%
    \begin{tabular}{l| c| cccc}
    \toprule
        & \textbf{Crop} & \multicolumn{4}{c}{\textbf{Weed}} \\
        Field Model & False Hits & Total Weeds & Acc. Hits & Par. Hits & Missed \\	 
    \midrule
    Weed-Only         & -   & 473 & 183 & 192 & 98  \\   
    \midrule
    Crop-Weed     & 4\% & 565 & 177 & 213 & 175 \\   
    \bottomrule
    \end{tabular}}
    \label{tab:realPerf}
\end{table}

Consequently, the weeding loss attributable to planning or intervention constraints stands at $11.6\%$.
This loss could be addressed by fine-tuning planning strategies, adding more weeding axes, increasing the velocity of the linear axis, or decreasing the robot's linear velocity.
The remaining $152$ weeds ($1038-886$) were missed because of systemic issues, notably the vision system's performance under difficult real-time and challenging lighting conditions. 
Hence, the overall loss of the system added up to $26.3\%$ of the total number of weed plants in the fields (including the detected-and-missed and not-detected instances). 

Further examination showed that in the weed-only region, from the total of $98$ missed weeds, a total of $39$ plants remained untreated solely because of the vision system's inability to detect them.
Similarly, in the crop-weed region, $113$ weeds were missed due to vision system failure highlighting the limitations of the vision system and the challenges encountered in real-field conditions.
This loss could be reduced by improving the vision system and using more advanced DNN architecture in the future.

In the real-world field trials of \bbot\ $765$ weeds were successfully treated.
Of the treated weeds, on average $47\%$ were considered to be treated in a highly accurate manner, $183$ and $177$ weeds in weed-only and crop-weed regions, respectively.
This means the spray footprint was centered on the weed.
The remaining $53\%$ were treated successfully but with less accuracy, with an offset of nearly $2cm$.
They are considered to be partial treatments as they still cover a large proportion of the weed.
This reduced performance is likely linked to the variable conditions encountered in real-field settings.
Factors such as slight weather changes, like unexpected wind gusts, can alter spray distribution by a few centimeters or move plant leaves, thereby impacting the accuracy of weed detection and treatment.
While these errors occurred due to several factors we still consider this to be a successful treatment of the weeds in the field.